\documentclass[letterpaper, 10 pt, conference]{ieeeconf}  

\IEEEoverridecommandlockouts                              

\usepackage{amsmath} 
\usepackage{booktabs}   
\usepackage{multirow}   
\usepackage{tabularx}   

\usepackage{cite}

\usepackage[utf8]{inputenc}
\usepackage[T1]{fontenc}
\usepackage{graphicx}

\usepackage{color}
\usepackage[table]{xcolor}
\usepackage{colortbl}

\definecolor{mygray}{gray}{.9}
\definecolor{ggray}{RGB}{127,127,127}
\definecolor{mygreen}{RGB}{93,174,86}
\definecolor{macaronBrown}{RGB}{205,170,125}

\definecolor{rblue}{rgb}{0,0.5,1}
\definecolor{awesome}{rgb}{1.0, 0.13, 0.32}
\definecolor{hollywoodcerise}{rgb}{0.96, 0.0, 0.63}
\definecolor{lasallegreen}{rgb}{0.03, 0.47, 0.19}
\definecolor{hanpurple}{rgb}{0.32, 0.09, 0.98}
\definecolor{green(pigment)}{rgb}{0.0, 0.65, 0.31}

\makeatletter
\let\NAT@parse\undefined
\makeatother
\usepackage[pagebackref=false, breaklinks=true, colorlinks, bookmarks=false]{
        hyperref
}
\hypersetup{
        colorlinks=true,
        linkcolor={red},
        citecolor={hanpurple},
        urlcolor={magenta}
}

\usepackage{amssymb}

\title{\LARGE \bf
EgoRefine: Ego-Referenced Predictive Alignment and Trajectory-Conditioned Reliability-Aware Fusion for Asynchronous Collaborative Perception}
\author{Lingzhao Kong$^{1}$, Yongsheng Zang$^{1}$, Yu Kang$^{2,3}$, Kailun Yang$^{2,3}$, Jie Fu$^{4}$, Yukun Zuo$^{2,3}$, and Zhiyong Li$^{1,2,3}$%
\thanks{This work was supported by the National Natural Science Foundation of China (Grant No. 62503163), in part by the Natural Science Foundation of Changsha (Grant No. kq2502039), and in part by the Scientific Research Project of the Hunan Provincial Department of Education (Key Project) (Grant No. 25A0052).
\textit{(Corresponding authors: Yukun Zuo and Zhiyong Li.)}}%
\thanks{$^{1}$L. Kong, Y. Zang, and Z. Li are with the School of Computer Science and Electronic Engineering, Hunan University, Changsha 410082, China.}%
\thanks{$^{2}$Y. Kang, K. Yang, Y. Zuo, and Z. Li are with the School of Artificial Intelligence and Robotics, Hunan University, Changsha 410012, China (email: yukunzuo@hnu.edu.cn; zhiyong.li@hnu.edu.cn).}%
\thanks{$^{3}$Y. Kang, K. Yang, Y. Zuo, and Z. Li are also with the National Engineering Research Center of Robot Visual Perception and Control Technology, Hunan University, Changsha 410082, China.}%
\thanks{$^{4}$J. Fu is with Nanjing University of Posts and Telecommunications, Nanjing 210003, China.}
}

\begin{document}

\maketitle
\thispagestyle{empty}
\pagestyle{empty}

\begin{abstract}
Collaborative perception enables connected agents to share complementary observations for 3D object detection, extending sensing range and mitigating occlusion. Under asynchronous communication, however, cooperative features arrive with temporal delay. Existing prediction-based methods compensate for these features mainly from the transmitting agent's own history, leaving residual misalignment with the ego agent's current observation; subsequent fusion also often overlooks spatial variations in alignment quality. We propose EgoRefine, an ego-referenced predictive alignment and reliability-aware fusion framework for asynchronous collaborative perception. Its Ego-referenced Predictive Alignment module uses the current ego feature to guide cooperative trajectory-field prediction and refines the sampling offsets along an ego-referenced trajectory direction. Its Trajectory-conditioned Reliability-aware Fusion module treats the trajectory discrepancy between the ego and cooperative streams and the directional refinement magnitude as alignment cues, using them to condition the relation between aligned features and adaptively reweight the two streams before convolutional fusion. Experiments on V2V4Real and DAIR-V2X-Seq show that EgoRefine outperforms TraF-Align by $1.6$ and $2.9$ points on average in AP@0.5 and AP@0.7, respectively. The source code will be made publicly available at \url{https://github.com/godk0509/EgoRefine}.
\end{abstract}

\section{Introduction}

Collaborative perception has emerged as a promising paradigm for autonomous driving, where connected vehicles and infrastructure agents share complementary perceptual information to extend the sensing range and mitigate occlusion~\cite{10517450, liu2023, Li_2023_ICCV,wang2023core,chen2023transiff}. In this work, we focus on collaborative 3D object detection, where an ego agent fuses observations from multiple connected agents to localize objects in dynamic traffic scenes. Existing systems commonly transmit intermediate Bird's-Eye-View (BEV) features from cooperative agents to the ego agent, which then aggregates them with its own feature for detection. However, such collaboration relies on wireless links that inevitably introduce latency in real-world vehicle-to-vehicle and vehicle-to-infrastructure scenarios~\cite{11077784,Zhang_2024_CVPR}, so that cooperative features received by the ego agent correspond to past timestamps. This is particularly harmful for moving objects, since their responses in a delayed feature are spatially displaced from the ego agent's current observation, and directly fusing them produces mislocalized boxes and duplicate detections. Asynchronous collaborative perception, which aims to exploit delayed cooperative features while maintaining consistency with the ego agent's current view, is therefore a practical and important problem.

\begin{figure}
    \centering
    \includegraphics[width=1.0\linewidth]{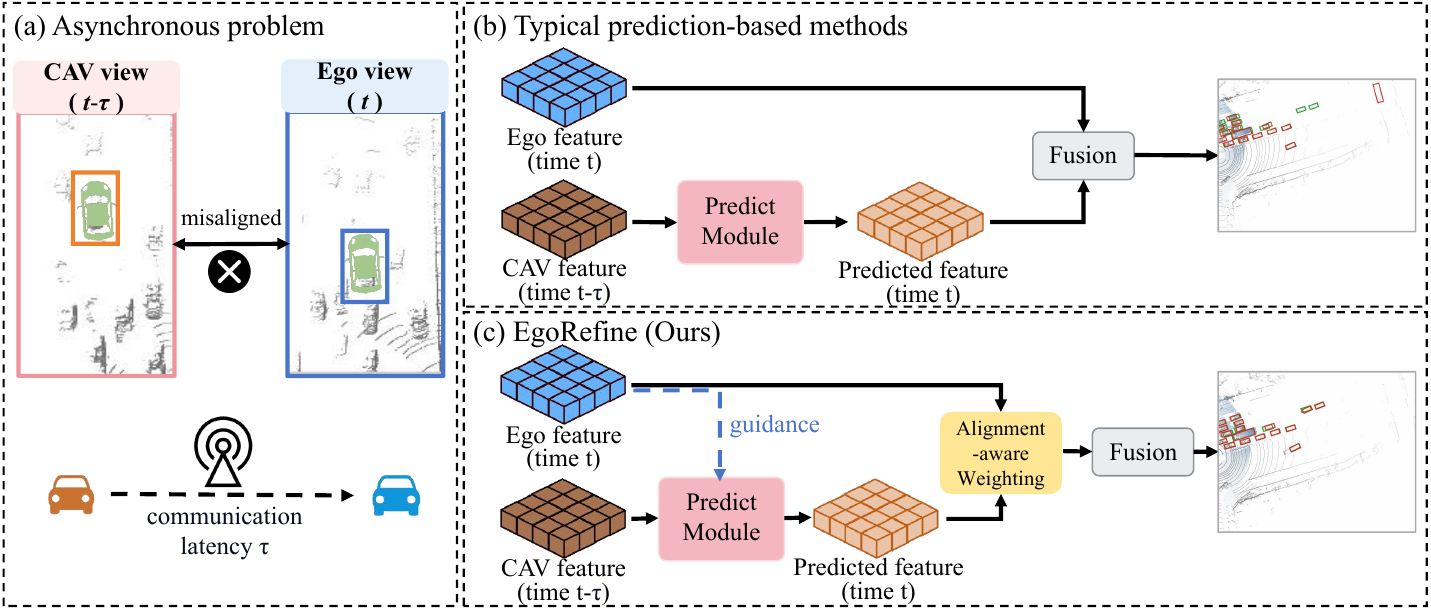}
    \vskip-1ex
    \caption{Motivation of EgoRefine. (a) Communication latency causes spatiotemporal misalignment between the delayed CAV observation and the current ego observation. (b) Typical prediction-based methods compensate for the delayed CAV feature from its own history and directly fuse it with the current ego feature. (c) EgoRefine uses the current ego feature to guide delayed-feature prediction and reweights the aligned ego and CAV features using alignment cues before fusion.}
    \label{fig:motivation}
    \vskip-3ex
\end{figure}

Compared with synchronous collaboration, temporal latency changes the spatial correspondence between agents. Throughout this paper, CAV denotes a cooperative agent, either a connected and automated vehicle or roadside infrastructure. Although a delayed CAV feature retains complementary cues, its object responses are shifted from their current positions by spatially varying amounts, making different regions differently reliable after alignment. Asynchronous collaboration therefore requires both predictive alignment and careful feature selection. Existing latency-compensation methods, including feature forecasting in SyncNet~\cite{lei2022syncnet} and FFNet~\cite{yu2023vehicle}, flow-based warping in CoBEVFlow~\cite{wei2023robust}, and trajectory-field prediction in TraF-Align~\cite{song2025traf}, predict the current-time cooperative feature mainly from the CAV's own history, without using the current ego representation to condition prediction. Such extrapolation cannot fully capture motion changes during the delay, leaving residual misalignment. Moreover, the concatenation-and-convolution fusion of TraF-Align and the attention-based~\cite{vaswani2017attention} fusion of V2X-ViT~\cite{xu2022v2xvit} and CoBEVT~\cite{xu2022cobevt} do not directly use alignment-stage cues to assess local reliability. Consequently, residually misaligned regions can contaminate the fused representation.

To address these issues, the key insight of this work is that the current ego feature should serve as a temporal and spatial reference for both predictive alignment and feature fusion, as shown in Fig.~\ref{fig:motivation}. For delayed CAV features, ego information should not simply be added as another feature source; instead, it should guide the prediction of how the delayed feature evolves toward the ego agent's current timestamp, since the ego feature provides direct current-time cues for co-visible objects. Furthermore, any correction to the predicted offset should follow the predicted motion direction rather than be learned as an independent, unconstrained offset, so that the refinement stays consistent with the underlying object motion. For feature fusion, the contributions of ego and aligned CAV features should be adaptively adjusted according to their local alignment status, rather than treating the compensated representation as uniformly reliable. Notably, the cross-agent discrepancy between the ego and ego-guided CAV trajectory fields and the magnitude of directional offset refinement provide complementary cues about local alignment status and can be reused for fusion without additional supervision.

Based on this insight, we propose EgoRefine, an ego-referenced predictive alignment and reliability-aware fusion framework for asynchronous collaborative perception. It comprises two modules: Ego-referenced Predictive Alignment (EPA) and Trajectory-conditioned Reliability-aware Fusion (TRF). EPA selectively injects the current ego feature into the delayed CAV feature before trajectory-field prediction, anchoring the predicted field to the current ego timestamp rather than the CAV's history alone. It then applies a scalar refinement along the predicted orientation field to the base offsets, preserving the trajectory-induced motion structure while limiting the correction degrees of freedom. TRF reuses the ego-CAV trajectory discrepancy and directional refinement magnitude as alignment cues. These cues condition the semantic relation between the aligned features, which is mapped to a spatial gate that adaptively reweights the ego and CAV representations before convolutional fusion. In this way, the current ego observation informs both where the delayed feature is reconstructed and how strongly it contributes to the final representation.

The main contributions of this paper are summarized as follows:
\begin{itemize}
\item We introduce an ego-referenced view for asynchronous collaborative perception, where the current ego feature serves as a temporal-spatial reference to guide both delayed CAV feature alignment and fusion, rather than relying primarily on the cooperative agent's own history for temporal compensation.
\item We propose EgoRefine, which consists of Ego-referenced Predictive Alignment that anchors prediction to the ego timestamp and refines offsets only along the predicted motion direction, and Trajectory-conditioned Reliability-aware Fusion that reuses ego-CAV trajectory discrepancy and directional refinement magnitude to condition reliability-aware feature fusion. 
\item Extensive experiments on two real-world datasets, V2V4Real and DAIR-V2X-Seq, show that EgoRefine outperforms TraF-Align under every latency from $0$ to $400$~ms, with average gains of $1.6/2.9$ points in AP@0.5/AP@0.7 and the largest margins under the stricter AP@0.7 criterion, while adding no extra supervision, annotation, or communication overhead.
\end{itemize}

\begin{figure*}[t]
  \centering
  \includegraphics[width=\linewidth]{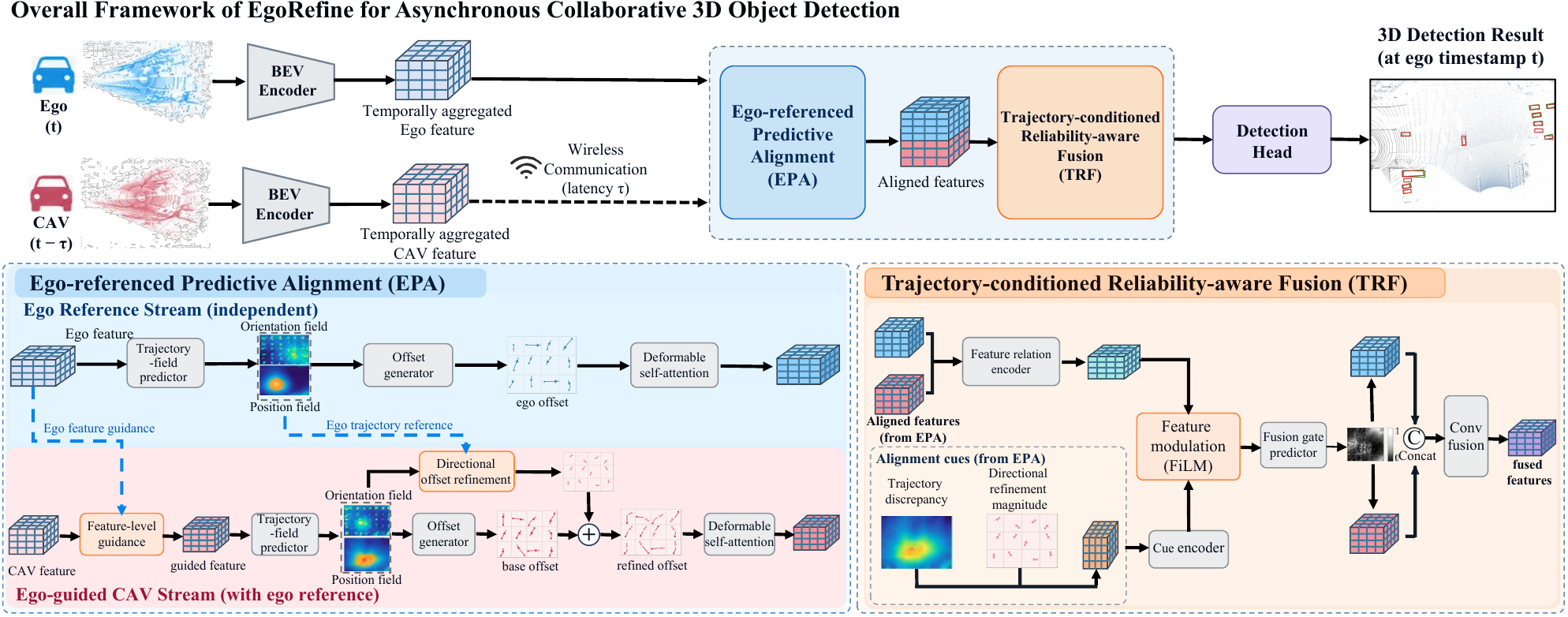}
  \vskip-1ex
  \caption{Overview of EgoRefine for asynchronous collaborative 3D object detection. EPA uses the current ego stream as a reference to guide delayed CAV features at both the feature and trajectory-offset levels, producing aligned ego and CAV representations. TRF further conditions the ego-CAV feature relation on ego-CAV trajectory discrepancy and directional refinement magnitude to estimate reliability-aware fusion weights before fusion and 3D detection.}
  \label{fig:framework}
  \vskip-3ex
\end{figure*}

\section{Related Work}
Our work is mainly related to two research directions: collaborative perception for 3D detection and asynchronous collaborative perception under communication latency.

\subsection{Collaborative Perception for 3D Object Detection}
Collaborative perception aims to improve autonomous driving perception by allowing connected vehicles and infrastructure agents to share complementary observations for 3D object detection~\cite{han2023collaborative,9676458,10801337,10802601,liu2026linking}. Existing methods can be broadly divided into early fusion, late fusion, and intermediate feature fusion~\cite{han2023collaborative}, among which feature-level collaboration has become dominant due to its balance between accuracy and communication cost. V2VNet~\cite{wang2020v2vnet} aggregates multi-agent features with graph neural networks, while DiscoNet~\cite{li2021learning} learns a distilled collaboration graph to improve communication efficiency. Transformer-based methods further enhance multi-agent interaction, where V2X-ViT~\cite{xu2022v2xvit} models V2X collaboration with spatial-aware attention and CoBEVT~\cite{xu2022cobevt} performs cooperative BEV fusion with sparse transformer blocks. Other methods focus on practical constraints in collaborative perception: CoAlign~\cite{lu2023robust} addresses pose errors, Where2comm~\cite{hu2022where2comm} selects spatially critical regions for communication, and HEAL~\cite{lu2024an} studies open heterogeneous collaborative perception. Recent works such as STAMP~\cite{gao2025stamp} and CoBEVMoE~\cite{kong2026cobevmoe} further improve scalability and heterogeneity modeling in multi-agent BEV perception. However, these methods mainly improve feature aggregation, communication efficiency, or heterogeneity under synchronized or near-synchronized settings, while temporal inconsistency caused by communication latency is not their primary focus. In contrast, our work focuses on asynchronous feature-level collaborative 3D detection, where delayed CAV features are aligned toward the ego agent's current timestamp and subsequently fused with reliability-aware adaptive weighting.

\subsection{Asynchronous Collaborative Perception}
Asynchronous collaborative perception studies how to exploit cooperative messages that arrive with temporal latency, so that delayed information can still benefit perception at the ego agent's current timestamp~\cite{yang2023how2comm,wang2025latency,xu2025codyntrust}. Existing approaches mainly investigate feature forecasting, motion-based compensation, and trajectory-aware alignment under latency. SyncNet~\cite{lei2022syncnet} synchronizes asynchronous features through feature-attention estimation and time modulation, while FFNet~\cite{yu2023vehicle} predicts future features for vehicle-infrastructure cooperative detection. CoBEVFlow~\cite{wei2023robust} explicitly estimates BEV flow to warp asynchronous cooperative features; How2comm~\cite{yang2023how2comm} adopts flow-guided delay compensation within a communication-efficient framework, and LRCP~\cite{wang2025latency} improves latency robustness by leveraging cached BEV features. These methods alleviate temporal misalignment through forecasting or warping, but their compensation is mainly driven by the transmitting agent's historical context. More recently, TraF-Align~\cite{song2025traf} predicts trajectory fields to guide deformable sampling offsets for temporal feature alignment, providing a strong trajectory-based solution for asynchronous multi-agent perception.
These methods perform temporal compensation and subsequent fusion as separate stages: compensation relies mainly on historical observations, while fusion does not directly use cues generated during alignment. EgoRefine instead couples alignment and fusion by using the current ego representation to guide CAV prediction and reusing the resulting ego-CAV trajectory discrepancy and directional refinement magnitude as alignment cues for feature fusion.

\section{Methodology}

\subsection{Problem Definition and Preliminaries}

Consider an ego agent and a cooperative agent. At the ego timestamp $t$, the ego agent maintains its current BEV representation, while the most recent feature received from the cooperative agent was captured at $t-\tau$, where $\tau$ denotes the communication latency. Following TraF-Align~\cite{song2025traf}, historical BEV features of each agent are augmented with temporal information, compensated for ego motion, and aggregated into temporally enriched representations. We denote the ego representation anchored at the current timestamp by $F_e\in\mathbb{R}^{C\times H\times W}$ and the delayed cooperative representation by $F_c\in\mathbb{R}^{C\times H\times W}$. The task is to predict object categories and bounding boxes at timestamp $t$, which requires compensating for the temporal discrepancy of $F_c$ before collaborative fusion.

Recent trajectory-based alignment methods follow a common prediction-and-reconstruction pipeline, which we adopt as our operating setting. A trajectory field $T=[T^{p},T^{o}]\in\mathbb{R}^{3\times H\times W}$ summarizes the motion tendency encoded by the temporally aggregated feature, where the one-channel position field $T^{p}$ marks the likely object locations along the trajectory and the two-channel orientation field $T^{o}$ encodes their motion directions toward the target timestamp. A field predictor $\Phi_{\mathrm{field}}(\cdot)$ predicts this trajectory field toward the ego timestamp, aiming to place ego and cooperative representations under a common temporal reference. An offset generator $\Phi_{\mathrm{off}}(\cdot)$ then converts the trajectory prior into $n$ deformable sampling displacements per location, $\Delta P\in\mathbb{R}^{n\times2\times H\times W}$, and a deformable self-attention operator $\mathcal{D}(\cdot,\cdot)$ gathers historical features according to these displacements to reconstruct the feature toward the target timestamp. Therefore, trajectory-field prediction determines how the scene evolves temporally, whereas offset generation translates this motion prior into spatial sampling locations for feature alignment.

Two limitations remain in this pipeline under communication latency. First, the delayed cooperative trajectory and offsets are predicted primarily from the transmitting agent's own historical information, making it difficult to account for motion changes occurring during $\tau$. Second, the intermediate alignment cues generated during predictive alignment are no longer explicitly exploited once the aligned feature is obtained, although the remaining alignment quality can vary substantially across spatial regions. To address these issues, we propose \textbf{EgoRefine}, consisting of \textbf{Ego-referenced Predictive Alignment (EPA)} and \textbf{Trajectory-conditioned Reliability-aware Fusion (TRF)}, as illustrated in Fig.~\ref{fig:framework}. Throughout this section, $\phi_{(\cdot)}$ denotes a lightweight convolutional network, $[\cdot]$ denotes channel-wise concatenation, $\odot$ denotes element-wise multiplication, and $\sigma(\cdot)$ denotes the sigmoid function.

\subsection{Ego-referenced Predictive Alignment}

The delayed cooperative feature alone cannot fully reveal motion changes occurring during communication latency, whereas the undelayed ego representation provides direct current-time cues for objects visible to both agents. EPA exploits this asymmetry through a dual-stream design. The ego stream is processed independently to preserve a clean current-time reference, while the cooperative stream receives ego guidance at both the feature-prediction and offset-refinement stages.

\textbf{Ego stream.}
Given the ego representation $F_e$, we first predict its trajectory field, sampling displacements, and reconstruct the ego representation:
\begin{equation}
T_e=\Phi_{\mathrm{field}}(F_e),\quad
\Delta P_e=\Phi_{\mathrm{off}}(T_e),\quad
F_e^a=\mathcal{D}(F_e,\Delta P_e).
\end{equation}
Although $F_e$ is anchored at the target timestamp and therefore requires no cross-agent temporal compensation, trajectory-guided feature reconstruction aggregates historical features along the predicted motion trajectories. The resulting $F_e^a$ is retained for collaborative fusion, while $T_e$ provides a current-time motion reference for the cooperative stream.

\textbf{Feature-level guidance.}
Directly injecting ego information everywhere may suppress the complementary observations preserved by the cooperative representation. We therefore adaptively regulate the ego contribution according to the local relation between the two agents:
\begin{equation}
G_t=\sigma\big(\phi_t([F_c,F_e,|F_c-F_e|])\big),
\end{equation}
where $|F_c-F_e|$ provides an explicit cue of local cross-agent discrepancy. Together with $F_c$ and $F_e$, it allows the guidance network to identify regions where current ego information can help calibrate the delayed cooperative representation. The ego-referenced cooperative feature is then obtained through a residual update:
\begin{equation}
\hat F_c=F_c+\alpha_t\,G_t\odot F_e,
\end{equation}
where $\alpha_t$ controls the feature-guidance strength. The residual formulation preserves the original cooperative representation, particularly for regions observed only by the CAV, while selectively introducing current-time ego cues.

\textbf{Ego-referenced field prediction.}
The guided cooperative feature is subsequently mapped to a trajectory field and corresponding base sampling displacements:
\begin{equation}
T_c=\Phi_{\mathrm{field}}(\hat F_c),\quad
\Delta P_c^{b}=\Phi_{\mathrm{off}}(T_c).
\end{equation}
Importantly, ego guidance is introduced before trajectory-field prediction rather than after feature reconstruction. Consequently, the field predictor can exploit current-time information while estimating how the delayed cooperative representation evolves toward timestamp $t$, reducing the dependence on history-only extrapolation. Errors already introduced into the trajectory prediction would otherwise be difficult to correct solely at the subsequent feature-fusion stage.

\textbf{Trajectory-directional refinement.} Although the ego-referenced field prediction improves the base displacements, residual offset errors may still remain. To preserve the motion structure encoded by the trajectory fields, we parameterize the additional correction along the predicted trajectory direction rather than learning another unconstrained two-dimensional displacement field. Specifically, a signed scalar refinement step is predicted for each sampling point:
\begin{equation}
S=s_{\max}\tanh\big(\phi_s([T_c,T_e,|T_c-T_e|])\big),
\end{equation}
where $S\in\mathbb{R}^{n\times H\times W}$ holds one step per sampling point and $s_{\max}$ bounds the refinement magnitude.

Following TraF-Align~\cite{song2025traf}, the orientation channels of the trajectory field are predicted through a sigmoid and thus lie in $[0,1]$; we first restore them to a signed domain:
\begin{equation}
\tilde d_j=2d_j-1,\qquad j\in\{e,c\},
\end{equation}
where $d_e,d_c\in\mathbb{R}^{2\times H\times W}$ denote the orientation components $T_e^{o}$ and $T_c^{o}$, respectively. Their ego-referenced motion direction is then obtained as
\begin{equation}
u=\operatorname{Norm}(\tilde d_c+\tilde d_e),
\end{equation}
where $\operatorname{Norm}(\cdot)$ denotes $\ell_2$ normalization along the channel dimension. The final cooperative sampling displacements are refined as
\begin{equation}
\Delta P_c=\Delta P_c^{b}+\alpha_d\,(S\otimes u),
\end{equation}
where $\otimes$ broadcasts each scalar step to the corresponding two-dimensional displacement and $\alpha_d$ controls the refinement strength. This structured parameterization reduces the residual prediction from $n$ unconstrained two-dimensional vectors to $n$ scalar steps, allowing the model to adjust the traveled distance along the predicted motion direction while avoiding unconstrained lateral corrections.

\textbf{Output.}
The ego-referenced cooperative representation is finally reconstructed as
\begin{equation}
F_c^a=\mathcal{D}(\hat F_c,\Delta P_c).
\end{equation}
EPA therefore provides TRF with not only the aligned representations $F_e^a$ and $F_c^a$, but also the intermediate trajectory fields $T_e$, $T_c$, and directional refinement step $S$, which record how the predictive alignment is obtained.

\begin{table*}[t]
\footnotesize
\centering
\caption{Performance comparison of the \textit{vehicle} class on the V2V4Real and DAIR-V2X-Seq validation sets under different communication latencies (AP@0.5/0.7, \%). Results of compared methods are taken from TraF-Align~\cite{song2025traf}. Best in bold, second best underlined.}
\setlength{\tabcolsep}{2.2pt} 
\renewcommand\arraystretch{1.1} 

\begin{tabular}{c|ccccc|ccccc}
    \toprule

    \multirow{2}{*}{{Method}} &\multicolumn{5}{c|}{{V2V4Real}} & \multicolumn{5}{c}{{DAIR-V2X-Seq}}  \\

    \cline{2-11}
    
    &{0 ms}&{100 ms}&{200 ms}&{300 ms}&{400 ms}&{0 ms}&{100 ms}&{200 ms}&{300 ms}&{400 ms}\\
    \hline  
    Single Agent &48.75/30.53&48.75/30.53&48.75/30.53&48.75/30.53&48.75/30.53&51.31/40.45&51.31/40.45&51.31/40.45&51.31/40.45&51.31/40.45 \\
    AttFuse \cite{xu2022opencood}&63.84/33.30&59.16/27.36&55.57/26.03&53.25/25.44&51.69/25.09&68.94/49.68&65.83/45.76&63.02/44.24&61.61/43.28&60.19/42.53\\
    F-Cooper \cite{chen2019f}&66.86/33.39&62.15/27.61&58.03/26.30&55.31/25.46&53.92/24.95&69.47/50.01&65.85/44.96&62.74/43.21&61.12/42.28&60.01/41.70\\
     V2VNet \cite{wang2020v2vnet} &67.11/35.47&61.97/29.14&58.09/27.85&56.26/27.43&54.62/26.75&71.44/51.91&67.26/46.63&63.84/44.59&61.68/43.24&59.78/42.08
\\    
      {Where2comm \cite{hu2022where2comm}}&67.03/37.55&61.86/31.52&58.22/30.20&56.22/29.71&54.47/29.07&68.37/48.50&64.81/44.32&61.51/42.80&59.66/41.63&58.25/40.80 \\
      V2X-ViT \cite{xu2022v2xvit} &68.51/35.88&63.45/29.57&59.42/28.04&57.35/27.65&55.60/27.20&70.21/50.92&66.42/46.44&62.58/44.29&61.00/43.56&59.63/42.66\\    
    MRCNet \cite{hong2024multi} &67.83/37.37&62.65/31.59&58.98/30.39&56.70/29.53&55.03/28.94&69.27/50.67&65.90/45.60&62.65/44.15&60.09/42.39&58.19/41.35\\
      CoBEVT \cite{xu2022cobevt} &70.59/38.82&65.50/32.11&61.55/30.40&58.91/29.54&57.05/29.26&71.51/51.31&67.60/46.55&64.38/45.03&62.39/44.11&60.74/43.08\\
      ERMVP \cite{Zhang_2024_CVPR} &70.74/38.45&65.53/31.91&61.63/30.53&59.24/30.18&57.26/29.31&71.19/{53.85}&67.04/48.09&63.67/46.37&61.23/44.76&59.55/43.93\\
\rowcolor{blue!10}
TraF-Align~\cite{song2025traf}
& \underline{74.28/44.14}
& \underline{73.04/39.99}
& \underline{71.51/39.54}
& \underline{70.07/38.99}
& \underline{69.41/38.49}
& \underline{76.90/58.31}
& \underline{73.95/53.68}
& \underline{72.69/52.83}
& \underline{71.66/52.33}
& \underline{71.22/52.01}\\
    
    \rowcolor{blue!20}
    {EgoRefine (Ours)}&\textbf{76.16/47.53}&\textbf{74.37/43.18}&\textbf{72.93/42.85}&\textbf{71.65/42.04}&\textbf{71.22/41.81}&\textbf{78.49/61.31}&\textbf{75.46/56.16}&\textbf{74.26/55.21}&\textbf{73.38/54.53}&\textbf{72.73/54.42}\\
    
\bottomrule
\end{tabular}

\label{tab:main_results}
\end{table*}

\subsection{Trajectory-conditioned Reliability-aware Fusion}

Predictive alignment substantially reduces temporal displacement, but the remaining alignment quality can still vary spatially. A natural starting point is to model the relation between the two aligned representations:
\begin{equation}
Z_f=
\phi_r([F_e^a,F_c^a,|F_e^a-F_c^a|]).
\end{equation}
The absolute feature difference exposes regions where the two aligned representations remain inconsistent. Such discrepancies alone are ambiguous, because they may result from residual temporal misalignment as well as legitimate cross-view complementarity caused by viewpoint, point density, or occlusion. TRF therefore conditions feature relation on motion-specific alignment cues produced during EPA.

\textbf{Alignment cues.}
We reuse two intermediate quantities naturally generated by EPA. The first is the ego-CAV trajectory discrepancy:
\begin{equation}
D_{\mathrm{traj}}=|T_e-T_c|\in\mathbb{R}^{3\times H\times W}.
\end{equation}
Compared with raw feature differences, $D_{\mathrm{traj}}$ provides a more motion-specific indicator of cross-agent alignment inconsistency because it directly compares the trajectory structures inferred by the two streams.

The second cue is derived from the directional refinement step $S$. Since $S$ contains one scalar correction for each sampling point, its magnitude is aggregated along the sampling-point dimension:
\begin{equation}
M_s=\rho(|S|)\in\mathbb{R}^{1\times H\times W},
\end{equation}
where $\rho(\cdot)$ averages over the $n$ sampling points. A larger $M_s$ indicates that the base offset requires stronger ego-referenced correction and serves as a complementary indicator of local alignment difficulty. The two cues are combined as
\begin{equation}
C_a=[D_{\mathrm{traj}},M_s]\in\mathbb{R}^{4\times H\times W}.
\end{equation}
Both are directly derived from intermediate quantities already produced by EPA and require no additional supervision.

\textbf{Alignment-conditioned modulation.}
The alignment cue $C_a$ is low-dimensional, whereas the relation
representation $Z_f$ contains high-dimensional semantic information. Directly concatenating them may dilute the alignment cues among semantic channels. We therefore treat $C_a$ as conditioning information in a feature-wise linear modulation manner~\cite{perez2018film}. A lightweight cue encoder predicts the modulation parameters:
\begin{equation}
[\gamma,\beta]=\phi_c(C_a),
\end{equation}
which modulate the feature-relation representation as
\begin{equation}
Z_m=Z_f\odot(1+\gamma)+\beta.
\end{equation}
This formulation allows the alignment cues to modulate how the ego-CAV feature relation is interpreted, rather than competing directly with semantic features through channel concatenation. Consequently, similar feature discrepancies can induce different fusion preferences according to the alignment status inferred from $D_{\mathrm{traj}}$ and $M_s$.

\textbf{Fusion gate.}
The modulated relation is mapped to a spatial fusion gate:
\begin{equation}
G_f=\sigma(\phi_g(Z_m)).
\end{equation}
We use a complementary weighting formulation because the desired decision is the relative contribution of ego and cooperative information at each spatial location:
\begin{equation}
F_e^w=G_f\odot F_e^a,\qquad
F_c^w=(1-G_f)\odot F_c^a.
\end{equation}
The complementary weights also maintain a bounded relative contribution between the two streams before convolutional aggregation. We emphasize that $G_f$ is learned end to end as a reliability-aware fusion preference rather than an explicitly calibrated reliability probability. The final fused representation is produced by
\begin{equation}
F_{\mathrm{fused}}=\Phi_{\mathrm{fuse}}\big([F_e^w,F_c^w]\big),
\end{equation}
where $\Phi_{\mathrm{fuse}}$ is a two-layer convolutional fusion operator.

\subsection{Joint Optimization of Alignment and Fusion}

EPA and TRF form a unified differentiable path from temporally misaligned inputs to collaborative predictions. EPA determines where delayed cooperative information is reconstructed, while TRF determines how the reconstructed information contributes to the fused representation. Both stages are optimized jointly with the collaborative detector, allowing alignment and fusion to co-adapt to the final detection task without introducing module-specific objectives.

\section{Experiments}

\subsection{Experimental Setup}

\noindent\textbf{Datasets.}
We evaluate EgoRefine on two real-world collaborative perception datasets, \textbf{V2V4Real}~\cite{xu2023v2v4real} and \textbf{DAIR-V2X-Seq}~\cite{yu2023v2xseq}. V2V4Real contains vehicle-to-vehicle sequences collected by two connected vehicles, while DAIR-V2X-Seq provides sequential vehicle-infrastructure observations at intersections; both support evaluation under temporal asynchrony. We follow the official data splits and evaluation protocols of both benchmarks for all experiments.

\noindent\textbf{Evaluation Metrics.}
We adopt average precision (AP) for 3D object detection as the evaluation metric. Specifically, we report \textbf{AP@0.5} and \textbf{AP@0.7}, where the IoU thresholds are set to $0.5$ and $0.7$, respectively. AP@0.5 reflects overall detection capability, while the stricter AP@0.7 criterion is more sensitive to localization accuracy and is therefore particularly informative for evaluating asynchronous feature alignment.

\noindent\textbf{Implementation Details.}
EgoRefine is built on the official TraF-Align codebase~\cite{song2025traf} and inherits its backbone, field predictor, offset generator, and trajectory-aware attention. 
Following TraF-Align, point clouds are discretized with a grid size of $0.4\times0.4$~m, the backbone output has one-fourth of the BEV resolution; $n=18$ sampling points are generated per location; two attention layers with four heads are used; two ego frames and four cooperative frames are used on V2V4Real, and four frames per agent on DAIR-V2X-Seq. 
A single model is trained per dataset with the latency $\tau$ uniformly sampled from $0$ to $400$~ms as augmentation, and evaluated at fixed $\tau\in\{0,100,200,300,400\}$~ms applied to the non-ego agent. We set $\alpha_t=0.2$, $\alpha_d=0.3$, and $s_{\max}=0.01$. 
All experiments run on a single NVIDIA RTX A6000 GPU for $60$ epochs with a batch size of $4$, using AdamW and a one-cycle learning-rate schedule.

\begin{table}[!t]
\footnotesize
\centering
\caption{Ablation study of EPA and TRF on DAIR-V2X-Seq under different communication latencies (AP@0.5/0.7, \%). The first row is the TraF-Align baseline.}
\setlength{\tabcolsep}{0.8pt} 
\renewcommand\arraystretch{1.3} 
\begin{tabular}{cc|c|c|c|c|c}
    \toprule
    \multirow{2}{*}{EPA}&\multirow{2}{*}{TRF}& \multicolumn{5}{c}{\cellcolor{mygray}{Time Latency} }  \\
    \cline{3-7}
    &&0 ms&100 ms&200 ms&300 ms&400 ms\\
    \midrule
     &&76.90/58.31&73.95/53.68&72.69/52.83&71.66/52.33&71.22/52.01\\ %
     $\checkmark$&&78.30/60.31&	75.32/55.48&	74.18/54.79&	73.30/54.17&72.21/53.43\\
$\checkmark$&$\checkmark$&\textbf{78.49/61.31}&\textbf{75.46/56.16}&\textbf{74.26/55.21}&\textbf{73.38/54.53}&\textbf{72.73/54.42}\\
\bottomrule

\end{tabular}
\label{tab:ablation_component}
\vskip-3ex
\end{table}

\subsection{Comparison with State of the Art}
Table~\ref{tab:main_results} compares EgoRefine with representative collaborative perception methods on V2V4Real and DAIR-V2X-Seq under communication latencies ranging from $0$ to $400$~ms. \textbf{EgoRefine consistently achieves the best AP@0.5 and AP@0.7 on both datasets across all evaluated latency settings}, demonstrating its effectiveness under both synchronous and asynchronous collaboration. Compared with the strongest baseline, TraF-Align~\cite{song2025traf}, EgoRefine improves AP@0.5/AP@0.7 by an average of $1.60/3.25$ points on V2V4Real and $1.58/2.49$ points on DAIR-V2X-Seq over the five latency settings. The gains are particularly pronounced under the stricter AP@0.7 criterion, indicating that EgoRefine mainly improves fine-grained localization rather than only increasing coarse detection recall.

These improvements can be attributed to the complementary designs of EPA and TRF. EPA uses the current ego representation as an explicit reference for delayed CAV prediction and further constrains offset refinement along the inferred trajectory direction, reducing motion-induced residual misalignment. TRF then exploits trajectory discrepancy and directional refinement magnitude as alignment cues to modulate the ego-CAV feature relation before fusion, thereby suppressing unreliable cooperative information. Notably, the advantage remains substantial under severe latency: at $400$~ms, EgoRefine outperforms TraF-Align by $1.81/3.32$ points on V2V4Real and $1.51/2.41$ points on DAIR-V2X-Seq in AP@0.5/AP@0.7. The gains are not confined to the asynchronous setting either: at $0$~ms, EgoRefine still improves AP@0.7 by $3.39$ and $3.00$ points on the two datasets, since ego-referenced alignment and reliability-aware fusion also correct cross-view inconsistency that exists even without latency. EgoRefine therefore improves ego-CAV feature consistency in general, with latency being the case where such inconsistency is most severe.

\begin{figure}[htbp]
  \centering
  \includegraphics[width=\linewidth]{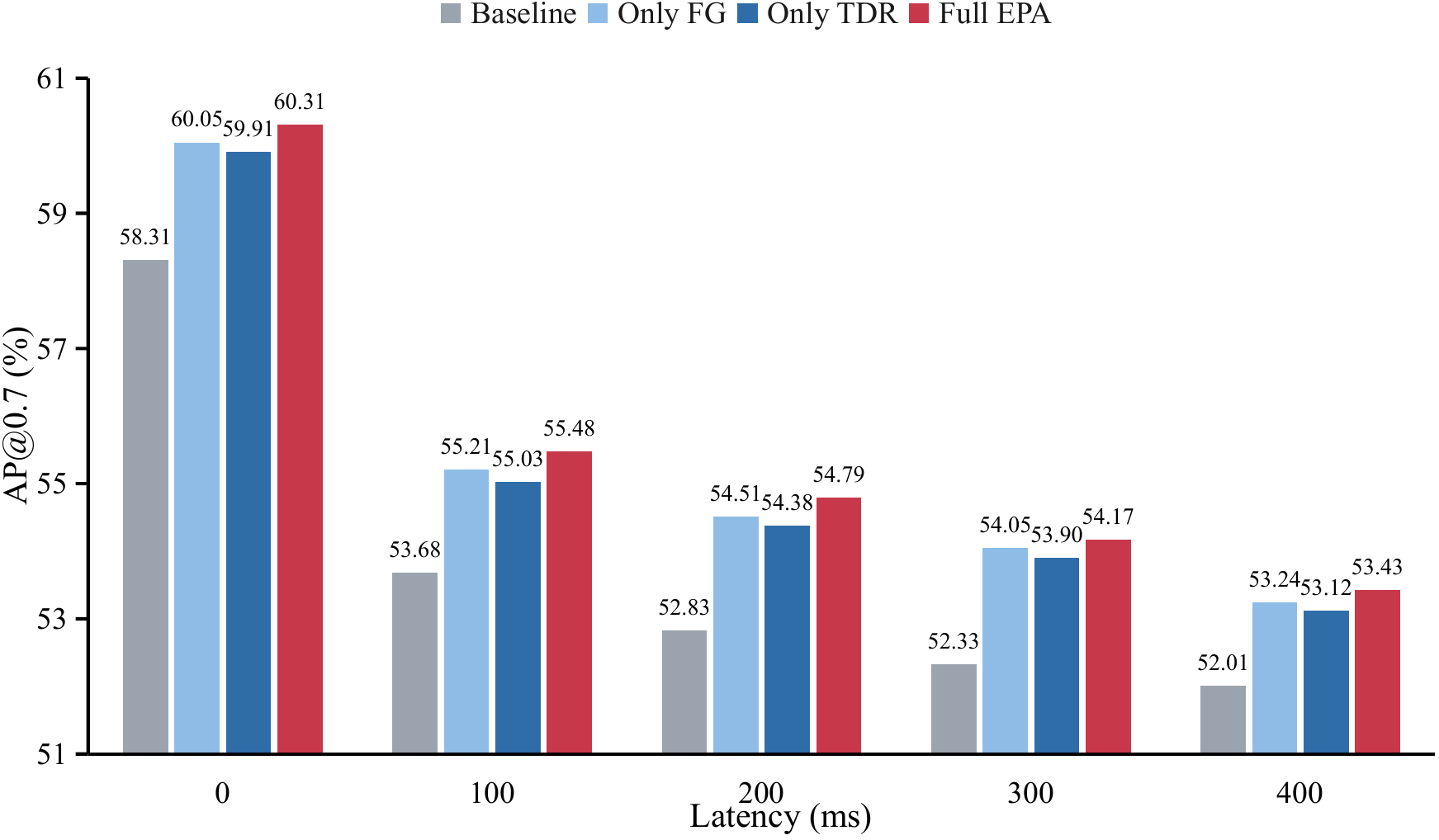}
  \caption{Ablation study of the two components in EPA on DAIR-V2X-Seq under different communication latencies. FG and TDR denote Feature-level Guidance and Trajectory-Directional Refinement, respectively.}
  \label{fig:ablation_EPA}
  \vskip-3ex
\end{figure}

\begin{figure*}[t]
  \centering
  \includegraphics[width=\linewidth]{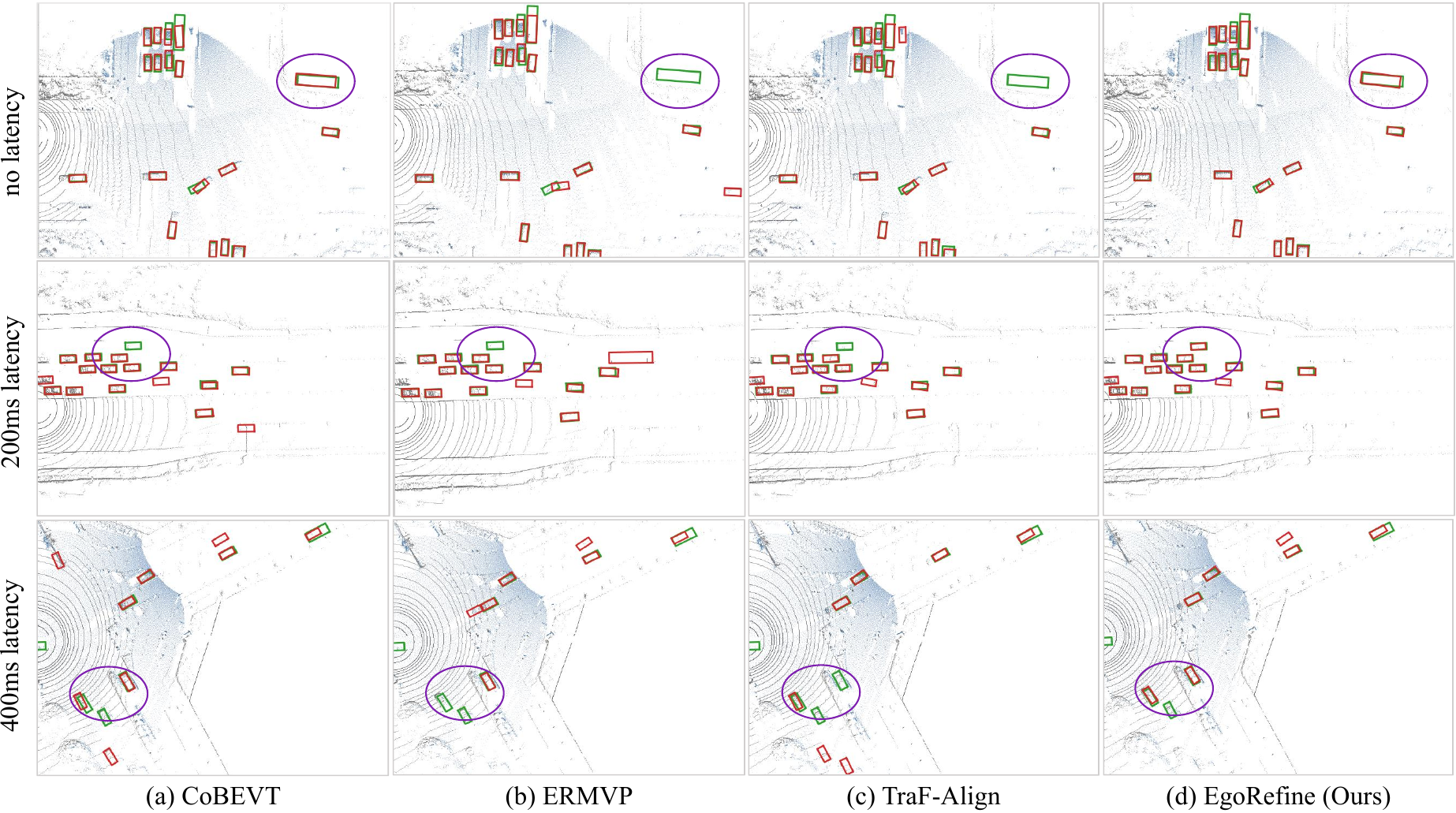}
  \vskip-1ex
  \caption{Qualitative comparison of CoBEVT, ERMVP, TraF-Align, and EgoRefine under different communication latencies. Green boxes denote ground-truth 3D bounding boxes and red boxes denote predicted results. The purple ellipses highlight representative regions where our method achieves more accurate localization and better robustness to temporal misalignment. In particular, EgoRefine preserves more stable detection quality as latency increases from 0 ms to 400 ms, while the competing methods exhibit more severe localization errors, missed detections, or inconsistent predictions.}
  \label{fig:vis_main}
  \vskip-3ex
\end{figure*}

\subsection{Ablation Studies}

We first evaluate the contributions of the two main components of EgoRefine, namely Ego-referenced Predictive Alignment (EPA) and Trajectory-conditioned Reliability-aware Fusion (TRF). As shown in Table~\ref{tab:ablation_component}, EPA consistently improves the TraF-Align baseline (first row) under all latency settings, yielding average gains of $1.38/1.80$ points in AP@0.5/AP@0.7. This verifies the effectiveness of introducing the current ego representation as a reference for delayed CAV prediction and trajectory-guided offset refinement. Adding TRF on top of EPA further improves AP@0.5/AP@0.7 by $0.20/0.69$ points on average, demonstrating that alignment-stage cues can effectively guide the subsequent fusion process. Overall, the complete EgoRefine achieves average improvements of $1.58/2.49$ points over the baseline, with the larger gain at AP@0.7 indicating a stronger benefit for fine-grained localization.

We further analyze the internal designs of EPA in Fig.~\ref{fig:ablation_EPA}. Feature-level Guidance (FG) alone improves AP@0.7 by $1.58$ points on average, confirming that the current ego feature provides useful temporal-spatial reference information for delayed CAV prediction. Trajectory-Directional Refinement (TDR) independently yields a $1.44$-point average improvement, showing that constraining offset correction along the predicted motion direction effectively reduces residual misalignment. Combining FG and TDR further increases the gain to $1.80$ points, demonstrating that feature-level ego guidance and trajectory-level offset refinement provide complementary benefits within EPA.

\begin{figure}[t]
  \centering
  \includegraphics[width=\linewidth]{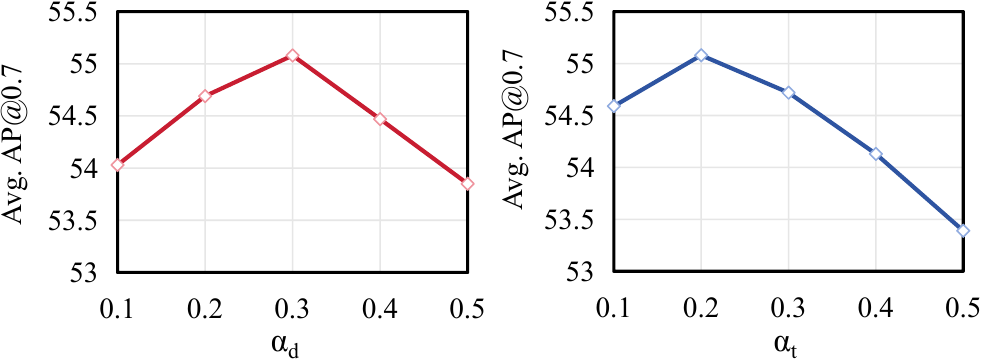}
  \caption{Hyperparameter sensitivity of the trajectory-directional refinement strength $\alpha_d$ and feature-guidance strength $\alpha_t$ on DAIR-V2X-Seq. AP@0.7 is averaged over the asynchronous latency settings from $100$ to $400$~ms, with all components of EgoRefine enabled.}
  \label{fig:hyp}
  \vskip-3ex
\end{figure}

\subsection{Hyperparameter Analysis}

We analyze the sensitivity of two key hyperparameters in EPA, namely the feature-guidance strength $\alpha_t$ and the trajectory-directional refinement strength $\alpha_d$. All components of EgoRefine are enabled, and only one parameter is varied at a time while the others are fixed. To reflect performance under asynchronous communication rather than a specific latency, we report AP@0.7 averaged over $100$--$400$~ms settings on DAIR-V2X-Seq, with $s_{\max}=0.01$ fixed throughout.

As shown in Fig.~\ref{fig:hyp}, the performance first improves and then decreases as either guidance strength increases. For $\alpha_t$, the best average AP@0.7 is obtained at $\alpha_t=0.2$, indicating that moderate ego feature guidance provides useful current-time information for delayed CAV prediction, whereas excessive injection may weaken the complementary information retained in the CAV representation. Similarly, $\alpha_d=0.3$ achieves the best performance for trajectory-directional refinement. A larger $\alpha_d$ gradually degrades detection accuracy, suggesting that overly strong directional correction may over-adjust the base offsets. We therefore set $\alpha_t=0.2$ and $\alpha_d=0.3$ in all experiments.

\subsection{Qualitative Results}

Fig.~\ref{fig:vis_main} presents qualitative comparisons of CoBEVT, ERMVP, TraF-Align, and EgoRefine under $0$, $200$, and $400$~ms communication latency. Overall, EgoRefine produces more accurate and stable 3D detection results across different latency settings. At $0$~ms, most methods already achieve reasonable detection quality, but EgoRefine still provides more precise localization in the highlighted regions, indicating that its ego-referenced alignment and reliability-aware fusion also improve feature interaction even in the synchronous setting.

As latency increases to $200$~ms and $400$~ms, the differences become more evident. The competing methods exhibit more pronounced localization errors, missing or duplicated boxes, and less stable predictions in crowded vehicle regions. In contrast, EgoRefine maintains better alignment with the ground-truth boxes, especially for the highlighted moving vehicles. This observation is consistent with the quantitative results in Tables~\ref{tab:main_results} and~\ref{tab:ablation_component}, where the proposed method yields larger gains under the stricter AP@0.7 metric. The qualitative examples therefore confirm that EgoRefine effectively alleviates temporal misalignment and improves localization robustness in asynchronous collaborative perception.

\section{Conclusions}
This work investigates the challenge of temporal misalignment and spatially varying alignment reliability in asynchronous collaborative 3D object detection. We propose \textbf{EgoRefine}, which combines ego-referenced predictive alignment with trajectory-conditioned reliability-aware fusion, and extensive experiments on V2V4Real and DAIR-V2X-Seq demonstrate consistent improvements over existing methods across different communication latencies. 
In future work, we plan to extend EgoRefine to more general multi-agent settings with multiple cooperative agents and dynamically varying communication conditions, where agent-wise alignment confidence and adaptive multi-agent fusion can be jointly modeled.

\begingroup
\renewcommand{\footnotesize}{\fontsize{8}{8.2pt}\selectfont}
\bibliographystyle{IEEEtran}
\bibliography{bibv1.bib}
\endgroup

\end{document}